\documentclass[journal,twoside,web]{IEEEtran}
\makeatletter
\let\NAT@parse\undefined
\makeatother

\usepackage{cite}
\usepackage{amsfonts}
\usepackage{graphicx}
\usepackage{textcomp}
\usepackage{graphicx}
\usepackage{caption}
\usepackage{multirow}
\usepackage{booktabs}

\usepackage{color}
\usepackage{empheq}
\usepackage{algorithm}
\usepackage{algpseudocode}
\usepackage{subcaption}
\usepackage{hyperref}
\usepackage{url}
\usepackage{mathtools, cuted}
\usepackage{lipsum, color}
\usepackage[pdftex,dvipsnames]{xcolor}  % Coloured text etc.
\usepackage{xcolor}

\usepackage{hyperref}
\usepackage{pifont}% http://ctan.org/pkg/pifont
\newcommand{\cmark}{\ding{51}}%
\newcommand{\xmark}{\ding{55}}%
\usepackage{amsmath}
\usepackage{amssymb}
\usepackage{bm}
\usepackage{mathtools}
\usepackage{dsfont}
\usepackage{multirow}
\usepackage{leftidx}
\usepackage{soul}
\usepackage{textgreek}

\DeclareMathOperator*{\argmin}{argmin}
\def\BibTeX{{\rm B\kern-.05em{\sc i\kern-.025em b}\kern-.08em
    T\kern-.1667em\lower.7ex\hbox{E}\kern-.125emX}}
\begin{document}

\title{Decoder-Only Image Registration}
\author{Xi Jia, Wenqi Lu, Xinxing Cheng, and Jinming Duan
\thanks{X. Jia, X. Cheng, and J. Duan are with the School of Computer Science, University of Birmingham, UK. W. Lu is with the Department of Computing and Mathematics, Manchester Metropolitan University, UK. The corresponding author is Jinming Duan (j.duan@cs.bham.ac.uk).}
}

\maketitle

\begin{abstract}

In unsupervised medical image registration, the predominant approaches involve the utilization of a encoder-decoder network architecture, allowing for precise prediction of dense, full-resolution displacement fields from given paired images. Despite its widespread use in the literature, we argue for the necessity of making both the encoder and decoder learnable in such an architecture. For this, we propose a novel network architecture, termed LessNet in this paper, which contains only a learnable decoder, while entirely omitting the utilization of a learnable encoder. LessNet substitutes the learnable encoder with simple, handcrafted features, eliminating the need to learn (optimize) network parameters in the encoder altogether. Consequently, this leads to a compact, efficient, and decoder-only architecture for 3D medical image registration. Evaluated on two publicly available brain MRI datasets, we demonstrate that our decoder-only LessNet can effectively and efficiently learn both dense displacement and diffeomorphic deformation fields in 3D. Furthermore, our decoder-only LessNet can achieve comparable registration performance to state-of-the-art methods such as VoxelMorph and TransMorph, while requiring significantly fewer computational resources. Our code and pre-trained models are available at this link \url{https://github.com/xi-jia/LessNet}

\end{abstract}


\section{Introduction}
\label{sec:introduction}

\IEEEPARstart{M}{edical} image registration aims to establish the spatial correspondence between a moving image and a fixed image. It plays an important role in diverse healthcare applications \cite{marti2001automatic,dean2012evaluation,hill2001medical}, including disease diagnosis, disease progression monitoring, treatment planning, treatment guidance, etc.

Traditionally, unsupervised medical registration has been addressed through iterative optimization approaches \cite{sotiras2013deformable}. These methods typically consist of three essential components. Firstly, a deformation model is defined, with options ranging from FFD \cite{rueckert1999nonrigid,rueckert2006diffeomorphic}, LDDMM \cite{beg2005computing}, DARTEL \cite{ashburner2007fast}, to Demons \cite{vercauteren2009diffeomorphic} and others \cite{sotiras2013deformable}. Secondly, an evaluation criterion is selected, which often incorporates a similarity constraint and a regularization term. Examples of similarity constraints include the mean squared error (MSE), normalized cross-correlation (NCC), mutual information (MI), and modality-independent neighborhood descriptor (MIND) \cite{heinrich2012mind}. Examples of regularization techniques encompass smoothness (e.g., diffusion regularizer, bending energy \cite{rudin1992nonlinear}, or arbitrary order gradient \cite{duan2023arbitrary}), inverse consistency \cite{christensen_2001consistent}, and diffeomorphism \cite{christensen1995topological}. Finally, the deformation model under the given criterion is iteratively optimized using a specialized optimization technique, such as the Levenberg–Marquardt algorithm \cite{ashburner2007fast}, deterministic or stochastic gradient descent\cite{rueckert1999nonrigid}, ADMM \cite{thorley2021nesterov}, etc. Such iterative registration approaches are designed to be explainable with clear mathematical derivations and can yield promising registration performance. A noteworthy aspect of these iterative approaches is their often slow optimization process and pairwise hyperparameter tuning, which restrict their application to large-scale imaging tasks. The emergence of recent data-driven approaches based on deep learning may have overcome these limitations \cite{Rueckert_Model_Data_Driven}.

\begin{figure}
    \centering
    \includegraphics[width=.45\textwidth]{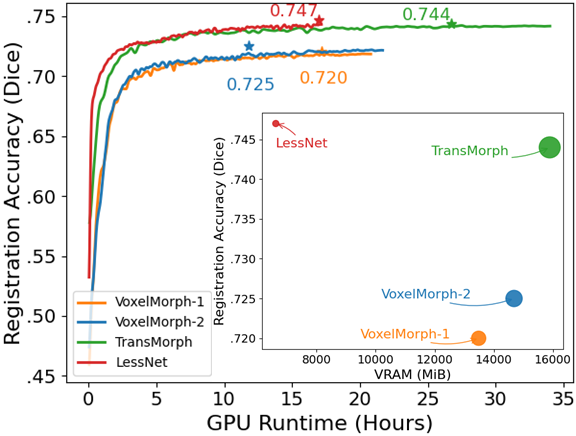}
    \caption{Comparison of registration performance, training time, memory usage, and mult-adds operations among different networks. In the main figure, $y$-axis represents registration accuracy measured by Dice on the validation set, while $x$-axis denotes GPU hours. Each network was trained for 500 epochs on a A100 GPU and our LessNet reached 500 epochs with the least amount of hours. In the subfigure, $y$-axis also represents registration accuracy, while $x$-axis denotes training memory usage measured in mebibytes (MiB). Within this subfigure, for each network, the area corresponds to the number of mult-adds required for one pair of images, each with a size of $160 \times 192 \times 224$. LessNet achieves superior registration accuracy with reduced training time, memory usage, and computational operations.}
    \label{fig:validation_accuracy_hours}
    \vspace{-10pt}
\end{figure}

In the context of deep learning-based unsupervised registration, neural networks replace conventional iterative processes and acquire registration knowledge through learning guided by an unsupervised loss function. Under this framework, there are two stages for registration: 1) training a network with a substantial amount of image pairs from a training set, and 2) deploying the trained network onto unseen image pairs from a test set to predict their displacement fields. More specifically, the registration is performed through a neural network ${f(\boldsymbol{\Theta})}$, parameterized by weights and biases $\boldsymbol{\Theta}$. Once trained, this network can efficiently predict a dense displacement field $\boldsymbol{u}= {{f}(\boldsymbol{\Theta}; I_M, I_F)}$ from the input image pair $I_M$ and $I_F$. The network enables fast inference with a single forward pass, resembling a closed-form solution without iteration, thereby outperforming the time-consuming nature of iterative optimization. For instance, VoxelMorph \cite{balakrishnan2018unsupervised} achieves comparable registration performance to iterative methods but runs orders of magnitude faster.

% resembling a closed-form solution without iteration

\begin{figure*}[t!]
    \centering
    \includegraphics[width=0.96\linewidth]{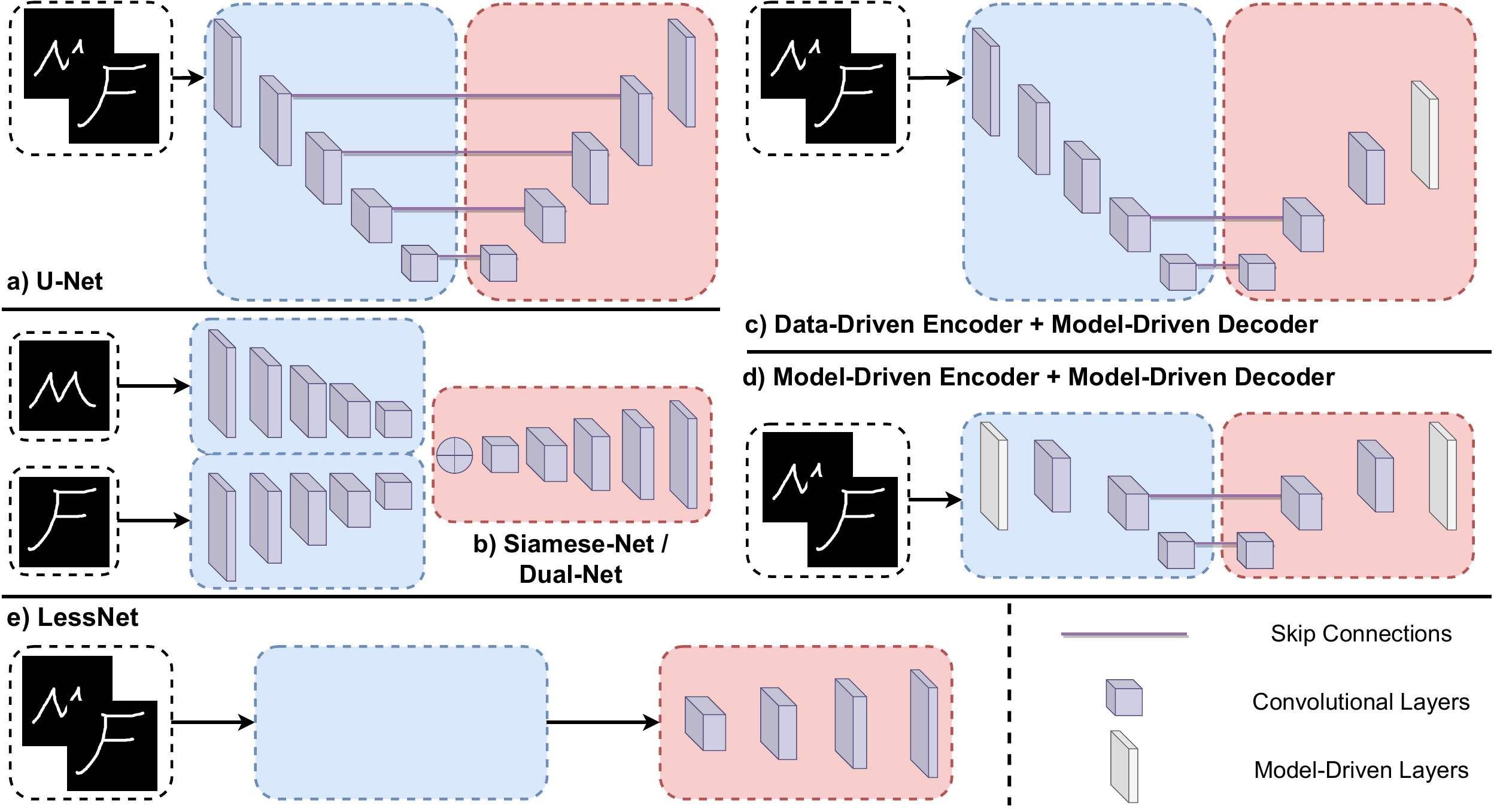}
    \caption{Semantic overview of different encoder-decoder architectures used in medical image registration. 
    a) U-Net style network: In this architecture, input moving and fixed images are combined into a two-channel image (marked by black box) as input. The encoder (marked by blue box) and decoder (marked by red box) exhibit a symmetric layout, and features from the encoder are skip-connected to those in the decoder. b) Siamese-Net (Dual-Net): In this architecture, the input moving and fixed images are fed into two parallel encoders, separately. Next, the decoder combines these features from the two encoders and then maps them to a registration field. 
    c) and d) Hybrid networks: In these architectures, some learnable layers are replaced by model-driven layers which are pre-defined, knowledge-driven parameter-free blocks. These hybrid approaches often lead to a reduced number of network parameters, fewer mult-adds, and thereby faster training and inference speeds. e) LessNet: Our network stands out by not having a learnable encoder at all. Instead, the decoder learns a full-resolution registration field directly from the input images.}
    \label{fig:OverviewArchitectures}
    \vspace{-10pt}
\end{figure*}

Following the success of VoxelMorph, which employs a U-Net architecture, numerous encoder-decoder style networks have been proposed for unsupervised image registration. Among them, some works, such as \cite{qin2018joint, qiustacom19, hu2019dual, KANG2022DualPRNet}, leverage siamese- or dual-style networks that incorporate two identical encoders to capture features from moving and fixed images, respectively. Another line of research, represented by works from \cite{Zhao_2019_ICCV, mok2020large, jia2021learning}, entails the progressive composition of intermediate displacement fields to form the final displacement field. These approaches involve either cascading multiple U-Nets or integrating multi-scale sub-layers into a U-Net style network. Very recently, some authors have explored vision-transformers to learn long-range information \cite{chen2021vit,chen2022transmorph}. While the debate between transformer- and convolution-based networks continues \cite{jia2022u}, a noticeable trend is the preference for larger networks with significantly more parameters and higher computational load. As an illustration, when comparing VoxelMorph to TransMorph, there is a respective increase of 136\%, 17046\%, and 216\% in memory usage, the number of parameters, and the number of mult-adds. Consequently, training such large networks often demands substantially longer time, increased computations, and extra GPU VRAM. Fig. \ref{fig:validation_accuracy_hours} further illustrates their detailed differences.

From VoxelMorph \cite{balakrishnan2018unsupervised,balakrishnan2019voxelmorph} and Siamese-Net or Dual-Net \cite{qin2018joint,qiustacom19,hu2019dual,KANG2022DualPRNet} to the more recent TransMorph \cite{chen2022transmorph}, the majority of registration networks have adopted the encoder-decoder (or contracting-expansion) style architecture. In this paradigm, input image pairs undergo a contracting path to encode high-level features, followed by an expansive path to decode these features into a dense, full-resolution displacement field. However, convolutions on full-resolution images or feature maps can entail intensive computations (mult-adds), especially when dealing with high-dimensional volumetric image data. Some works, such as DeepFlash \cite{zhang2019fast}, B-Spline \cite{de2019DLIR,qiu2021learning}, Fourier-Net \cite{jia2023fourier,jia2023fourier}, have recognized this drawback and proposed approaches to learn low-dimensional representations of the displacement field, which can significantly reduce the computational load resultant from convolutional operations in either the encoder or the decoder. Unfortunately, these networks exhibit limitations, as discussed in Sec. \ref{sec:related_works}.

To handle high-dimensional volumetric image data more effectively, we propose LessNet, which eliminates the entire learnable encoder and relies solely on a convolutional decoder to learn displacement fields from image pairs. A semantic comparison between the architectures of some popular registration networks as well as our LessNet is given in Fig. \ref{fig:OverviewArchitectures}. The main contributions of this work are summarized as follows:
\begin{itemize}
    \item While the majority of deep unsupervised registration networks adopted symmetric encoder-decoder (or contracting-expansion) style architectures, we demonstrated the presence of redundancy in the encoder and thereby proposed a simplified decoder-only architecture for medical image registration.
    \item As a proof of concept, we employed simple handcrafted features to replace the entire trainable encoder. These features comprise three distinct pooling operations, namely max pooling, average pooling, and min pooling. We showed in our experiments that these manually designed features are already effective for the decoder to learn dense, full-resolution displacement fields from image pairs. It is important to highlight that, beyond these pooling features, alternative choices such as incorporating other handcrafted features or features from large pre-trained networks may also be valid options.
    \item Evaluated on two brain MRI datasets for inter-subject and atlas-to-subject registration tasks, our decoder-only LessNet demonstrated comparable accuracy to state-of-the-art encoder-decoder style networks, while reducing computational load significantly, as illustrated in Fig. \ref{fig:validation_accuracy_hours}.
\end{itemize}

\section{Related Works}
\label{sec:related_works}
\subsection{Encoder-Decoder Style Networks}
Early efforts have been made to estimate dense displacement fields for 3D deformable image registration, as demonstrated in \cite{li2018non}, where the authors introduced a voxel-to-voxel encoder-decoder style fully convolutional network (FCN). Such an FCN model was optimized using Adam, with NCC as the data term and total variation as the regularization term. Balakrishnan et al. \cite{balakrishnan2018unsupervised, balakrishnan2019voxelmorph} further advanced the field by introducing VoxelMorph for unsupervised deformable image registration, which employs U-Net \cite{ronneberger2015u} as its backbone instead of the FCN \cite{li2018non}. In this framework, they explored the use of MSE and local normalized cross-correlation (LNCC) as the data term, along with a diffusion regularizer. Additionally, in \cite{balakrishnan2019voxelmorph} a Dice loss based on anatomical segmentation masks was incorporated during training to improve registration accuracy.

Qin et al. \cite{qin2018joint} presented a framework for jointly learning motion and segmentation in cardiac sequences. They utilized a Siamese-style recurrent network as the backbone, with the loss function incorporating the MSE between warped frames and the target frame as well as the regularization that induces both spatial and temporal smoothness of the displacement fields. Hu et al. introduced the dual-stream pyramid network (Dual-PRNet) \cite{hu2019dual} and later extended it to Dual-PRNet${++}$ \cite{KANG2022DualPRNet}. In contrast to VoxelMorph, which is a single-stream encoder-decoder network that predicts the full-resolution displacement field from the last convolutional layer, Dual-PRNet predicts multiple intermediate displacement fields in a coarse-to-fine manner. The final displacement field is integrated through upsamplings and warpings on intermediate displacement fields. Their loss function includes the NCC and diffusion regularizer.

Zhao et al. \cite{Zhao_2019_ICCV} introduced a recursive cascaded network architecture, which they termed RC-Net. This approach enables sequential warpings of the moving image, facilitating the estimation of large displacement fields. Within RC-Net, each cascade employs a U-Net style network to predict a small displacement field for an image pair, which consists of the warped moving image from the previous cascade and the original fixed image. The final displacement field is a composite of all small displacement fields predicted by each cascade. Using multiple cascaded networks (e.g., 10 cascades in their study), RC-Net demonstrated substantial improvement over VoxelMorph \cite{balakrishnan2018unsupervised} in both liver and brain registration tasks. Jia et al. proposed VR-Net \cite{jia2021learning}, which unrolls the mathematical structure of an iterative variational optimization through variable splitting and seamlessly integrating it into a deep neural network in a cascading fashion. However, optimal performance in VR-Net relies on the existence of an initial displacement field, which is predicted by an additional U-Net.

While algorithms such as \cite{balakrishnan2018unsupervised,balakrishnan2019voxelmorph,de2019DLIR,Zhao_2019_ICCV,jia2021learning} demonstrated efficacy in fast unsupervised registration, they lack a guarantee or promotion of inverse-consistency and topology preservation during the registration process. To address these limitations, Zhang \cite{zhang2018inverse} proposed an inverse-consistent U-Net style network (IC-Net), which enforces inverse consistency on both the forward and backward displacement fields. This ensures that the image pair is symmetrically deformed towards each other. To further mitigate folding issues in the deformation, they introduced an anti-folding constraint along with a local smoothness term. In the work by Dalca et al. \cite{dalca2018unsupervised} and its subsequent extension \cite{dalca2019unsupervised}, a probabilistic diffeomorphic registration network was introduced to learn diffeomorphisms for deformation fields. Initially, they proposed to learn a distribution of stationary velocity fields using a variational 3D U-Net architecture. Diffeomorphism was achieved by applying scaling and squaring \cite{ashburner2007fast} to the stationary velocity field. Building upon IC-Net and the probabilistic diffeomorphic registration networks \cite{zhang2018inverse,dalca2018unsupervised}, several works have been introduced to further enhance the accuracy of diffeomorphic registration, including SYM-Net \cite{Mok_2020_CVPR}, LapIRN \cite{mok2020large}, and CycleMorph \cite{kim2021cyclemorph}.

Recently, architectures based on vision transformers \cite{dosovitskiy2020image} have drawn a lot of attention in the registration community due to their capacity to capture long-range dependencies \cite{chen2021vit,chen2022transmorph}. It is noteworthy that despite the replacement of basic convolutional blocks with more advanced attention blocks in these transformer-based architectures, they still follow the classical encoder-decoder style \cite{jia2022u}.

\subsection{Model-Driven Networks}
Instead of making an entire network learnable across all layers, model-driven networks replace some learnable layers with pre-defined, knowledge-driven parameter-free modules. This approach often leads to a reduced number of parameters, fewer mult-adds, and thereby faster training and inference speeds \cite{de2017end,de2019DLIR,qiu2021learning,zhang2019fast,jia2023fourier,jia2023fourier+}. 

B-Spline networks such as DIR-Net \cite{de2017end} and DLIR \cite{de2019DLIR} estimated a grid of control points. The full-resolution displacement field was then interpolated from these points using a cubic b-spline function, which serves as a mathematical model. Qiu et al. \cite{qiu2021learning} introduced Diff-B-Spline, a diffeomorphic b-spline network designed for modality-invariant registration utilizing mutual information. In \cite{qiu2021learning}, they pruned the decoder by discarding several convolutional layers to predict b-spline parameterized velocities. These velocities, represented as regularly spaced control points, can then be interpolated into dense diffeomorphic deformation fields using scaling and squaring. B-Spline networks \cite{de2017end,de2019DLIR,qiu2021learning} typically require a local interpolation process. This characteristic often leads to a compromise in capturing global details, consequently impacting their overall registration performance.

The multi-step DeepFlash \cite{zhang2019fast} performs registration by first predicting a band-limited velocity field and subsequently converting it to the full-resolution deformation field using a model-driven partial differential equation. Note that the multi-step process makes DeepFlash cumbersome and difficult to implement. It is also important to note that DeepFlash relies on supervision signals for training, and therefore its performance may be bounded by the effectiveness of the underlying iterative method \cite{zhang2019fast} used to generate these supervision signals.

Fourier-Net \cite{jia2023fourier} is an end-to-end and unsupervised approach capable of predicting a compact, low-dimensional representation of the displacement field in the band-limited Fourier space. Within Fourier-Net, a model-driven decoder effectively reconstructed the full-resolution displacement field from a few band-limited coefficients. Building upon Fourier-Net, the authors introduced Fourier-Net+ \cite{jia2023fourier+}, which learns the band-limited displacement field from the band-limited representation of images. Fourier-Net+ further accelerated registration speed by constraining both the input and output of the network to low-dimensional representations, thereby reducing the need for repeated convolution operations. However, the process of learning the low-dimensional band-limited representation often results in the loss of high-frequency signals in displacement fields. To address this limitation, the use of multiple cascades, as demonstrated in Fourier-Net+ \cite{jia2023fourier+}, became necessary to effectively handle complex and large displacement fields.

\section{LessNet}

\subsection{Redundancy in Encoder}
The use of encoder-decoder architectures has been playing an essential role in medical image registration. For this, we ask a question: do the encoder and the decoder equally contribute to the estimation of displacement fields? This question has not been answered by researchers in the field of image registration using deep learning. To answer it, we used the classical architecture of VoxelMorph as an example. More specifically, we employed VoxelMorph-1 in \cite{balakrishnan2018unsupervised} and VoxelMorph-2 in \cite{balakrishnan2019voxelmorph} (their details have been given described in Section \ref{sec:exp}) and conducted the following experiments for each network:
\begin{itemize}
    \item We randomly initialized all parameters in the encoder and the decoder and trained these parameters from scratch.
    \item We randomly initialized all parameters in the encoder and the decoder, then freezed the decoder parameters (except the final output layer), and trained the encoder parameters (and the final output layer) only.
    \item We randomly initialized all parameters in the encoder and the decoder, then freezed the encoder parameters and trained the decoder parameters only.
\end{itemize}

Analysis of the results in Table \ref{tab:vm_encoder_decoder_only} indicates that the learning of the encoder is less critical than the learning of the decoder. Consequently, we argue that, for medical image registration, a learnable encoder may not be necessary if the decoder is appropriately designed. Furthermore, we note that even a randomly initialized encoder in Table \ref{tab:vm_encoder_decoder_only} can still incur considerable convolutional computations and memory usage. This raises the question: can handcrafted features be employed as an alternative? The answer is Yes. 

\begin{table}[t]
\caption{The implications of enabling or disabling learning in the encoder or decoder of VoxelMorph-1 and VoxelMorph-2. \cmark\;signifies the activation of learning, while \xmark\;indicates the deactivation of learning.}
\centering
\label{tab:vm_encoder_decoder_only}
\begin{tabular}{lcccc}\hline
Methods     & Encoder & Decoder & Dice$\uparrow$        & $|J|_{ < 0}\%$  \\\hline
Initial       & - & - & 0.544$\pm$0.089 & -\\
VoxMorph-1 &  \cmark & \xmark & 0.679$\pm$0.051 & 0.486$\pm$0.341 \\
VoxMorph-2 &  \cmark & \xmark & 0.690$\pm$0.051 & 0.537$\pm$0.350 \\\hline
VoxMorph-1 &  \xmark & \cmark & 0.747$\pm$0.043 & 0.818$\pm$0.400 \\
VoxMorph-2 &  \xmark & \cmark & 0.747$\pm$0.043 & 0.826$\pm$0.390 \\\hline
VoxMorph-1 &  \cmark & \cmark & 0.757$\pm$0.039 & 0.723$\pm$0.370 \\
VoxMorph-2 &  \cmark & \cmark & 0.757$\pm$0.040 & 0.793$\pm$0.405 \\\hline
\end{tabular}
\end{table}

\begin{figure*}[t!]
    \centering    \includegraphics[width=0.96\linewidth]{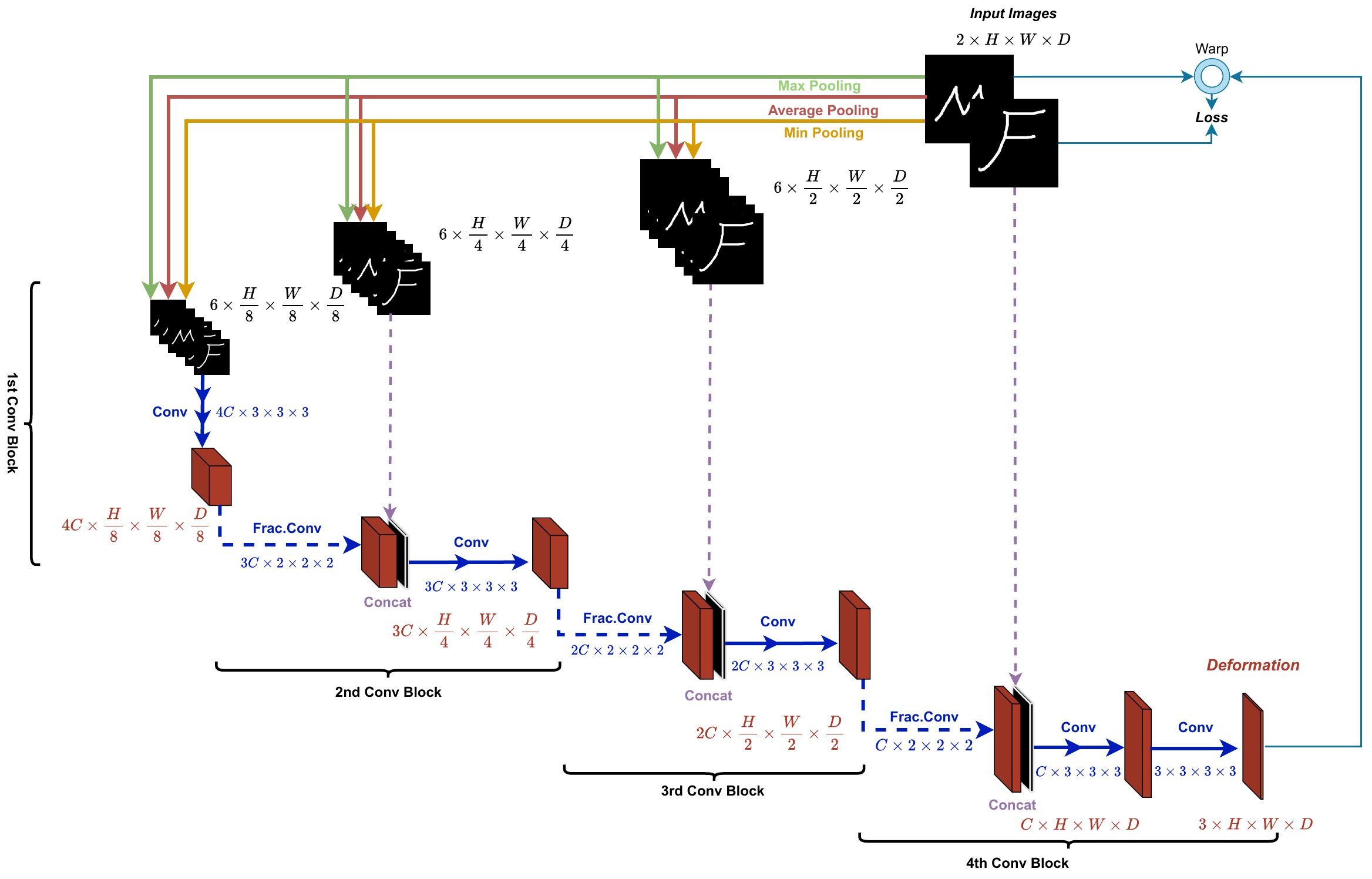}
    \vspace{-5pt}
    \caption{The architecture of LessNet. The upper half panel demonstrates the generation of multi-scale pooling features, while the lower half panel showcases the input and output of the learnable decoder, which consists of four hierarchical convolutional blocks. The loss function is applied to the moving image, warped by the predicted displacement field, and the fixed image.}
    \label{fig:LessNetArchitecture}
    \vspace{-10pt}
\end{figure*}

\subsection{Pooling Features}
\label{subsec:method_pooling}
As a proof of concept, we employed three parameters-free, computationally efficient pooling operations to substitute learnable layers in the conventional encoder. As illustrated in Fig. \ref{fig:LessNetArchitecture}, the original 3D image pair comes with a resolution of $2 \times H \times W \times D$. Employing a max pooling operation with a kernel size of $2 \times 2 \times 2$ and a stride of 2, a feature map of dimensions $2 \times \frac{H}{2} \times \frac{W}{2} \times \frac{D}{2}$ was generated. Similarly, using average pooling and minimum pooling operations, we obtained corresponding feature maps. By concatenating these pooling features, we created a six-channel feature map with a spatial resolution of $\frac{H}{2} \times \frac{W}{2} \times \frac{D}{2}$ in the first layer. Next, employing a kernel size of $4 \times 4 \times 4$ with a stride of 4 on the original input pair, a six-channel feature map was obtained with a spatial resolution of $\frac{H}{4} \times \frac{W}{4} \times \frac{D}{4}$ in the second layer. Subsequently, utilizing a kernel size of $8 \times 8 \times 8$ with a stride of 8 on the original input pair, a six-channel feature map was generated with a spatial resolution of $\frac{H}{8} \times \frac{W}{8} \times \frac{D}{8}$ in the third layer.

The $6 \times \frac{H}{8} \times \frac{W}{8} \times \frac{D}{8}$ pooling feature map was used as input to the first decoding layer. Simultaneously, the remaining pooling features, such as $6 \times \frac{H}{4} \times \frac{W}{4} \times \frac{D}{4}$ and $6 \times \frac{H}{2} \times \frac{W}{2} \times \frac{D}{2}$, were concatenated into their corresponding decoder layers as supplementary cues for registration. Additionally, the original image pairs was inserted into an appropriate decoding layer to provide additional information.

In these instances, we effectively leveraged the information contained in the moving and fixed image pair without the necessity of a learnable encoder. It is important to mention that alternative handcrafted features or features extracted from pre-trained networks are also valid options. We encourage the reader to experiment with various handcrafted or pre-trained features that could potentially improve registration performance. However, for the purpose of this study, we limited our exploration to simple pooling features to showcase the feasibility of decoder-only architectures for image registration.

\subsection{Decoder in LessNet}
As shown in Fig. \ref{fig:LessNetArchitecture}, the decoder in LessNet consists of four hierarchical convolutional blocks. In the following, we provide a detailed explanation of each convolutional block:

\begin{itemize}
    \item \textbf{The first convolutional block}: The input to this block is a six-channel feature map with a spatial resolution of $\frac{H}{8} \times \frac{W}{8} \times \frac{D}{8}$. This feature map undergoes convolution with kernels of size $4C \times 3 \times 3 \times 3$. Finally, the output of this block is a $4C \times \frac{H}{8} \times \frac{W}{8} \times \frac{D}{8}$ feature map.
    \item \textbf{The second convolutional block}: The input to this block is the $4C \times \frac{H}{8} \times \frac{W}{8} \times \frac{D}{8}$ feature map from the last block. This feature map is firstly upsampled to $3C \times \frac{H}{4} \times \frac{W}{4} \times \frac{D}{4}$ using fractional convolution with kernels of size $3C \times 2 \times 2 \times 2$. The upsampled feature map is subsequently concatenated with the $6 \times \frac{H}{4} \times \frac{W}{4} \times \frac{D}{4}$ pooling features, yielding a new feature map of size $ (3C+6) \times \frac{H}{4} \times \frac{W}{4} \times \frac{D}{4}$. This new feature map undergoes convolution with kernels of size $3C \times 3 \times 3 \times 3$. As a result, the output of this block is a $3C \times \frac{H}{4} \times \frac{W}{4} \times \frac{D}{4}$ feature map.
    \item \textbf{The third convolutional block}: The input to this block is the $3C \times \frac{H}{4} \times \frac{W}{4} \times \frac{D}{4}$ feature map from the last block. This feature map is firstly upsampled to $2C \times \frac{H}{2} \times \frac{W}{2} \times \frac{D}{2}$ using fractional convolution with kernels of size $2C \times 2 \times 2 \times 2$. The upsampled feature map is subsequently concatenated with the $6 \times \frac{H}{2} \times \frac{W}{2} \times \frac{D}{2}$ pooling features, yielding a new feature map of size $ (2C+6) \times \frac{H}{2} \times \frac{W}{2} \times \frac{D}{2}$. Next, this new feature map undergoes convolution with kernels of size $2C \times 3 \times 3 \times 3$. As a result, the output of this block is a $2C \times \frac{H}{2} \times \frac{W}{2} \times \frac{D}{2}$ feature map.
    \item \textbf{The fourth convolutional block}: The input to this block is the $2C \times \frac{H}{2} \times \frac{W}{2} \times \frac{D}{2}$ feature from the last block. This feature map is firstly upsampled to $C \times H \times W \times D$ using fractional convolution with kernels of size $C \times 2 \times 2 \times 2$. The upsampled feature map is then concatenated with the original image pair, yielding a new feature map of size $ (C+2) \times  H \times W \times D$. Next, this new feature map undergoes convolution with kernels of size $C \times 3 \times 3 \times 3$, producing yet another feature map of size $C \times H \times W \times D$. This feature map then undergoes a final convolution with kernels of size $3 \times 3 \times 3 \times 3$, outputting the dense displacement field of size $3 \times H \times W \times D$. 
\end{itemize}
Note that, we used LeaklyReLU with a default negative slope of 0.01 as the activation function after each convolution apart from the final output layer, for which we used the SoftSign activation function. On the other hand, we crafted feature maps of size $\frac{H}{h} \times \frac{W}{w} \times \frac{D}{d}$ throughout the entire network, where $h=w=d$, and their values were either 2, 4, or 8. However, the values of $h$, $w$, and $d$ do not necessarily have to be equal, and one can choose appropriate values with slight modifications to the proposed architecture in Fig. \ref{fig:LessNetArchitecture}.

\subsection{Loss Functions}
The proposed LessNet is a general unsupervised learning framework, making it adaptable to typical registration loss functions. First, we assume that LessNet predicts directly displacement fields. In this case, the loss function ${\mathcal L}({\boldsymbol{\Theta}})$ for LessNet is given by:
\begin{equation}\label{eq:1}
\argmin_{\boldsymbol{\Theta}}  \frac{1}{N} \sum_{i=1}^N {\mathcal L}_{S}( I_{M_i} \circ  ({\boldsymbol{u}_i}+{\rm{Id}}), I_{F_i} ) + \frac{\lambda}{N} \sum_{i=1}^N {\mathcal L}_{R}({\boldsymbol{u}_i}),  
\end{equation}
with 
\begin{equation} \nonumber
    {\boldsymbol{u}_i} = {f}(\boldsymbol{\Theta}; I_{M_i}, I_{F_i}).
\end{equation}
Here, ${\boldsymbol{u}_i}$ denotes the displacement field, predicted by the network $f$ parameterized by ${\boldsymbol{\Theta}}$ on the input image pair $I_{M_i}$ and $I_{F_i}$. ${\rm{Id}}$ is the identity grid, $N$ is the number of training pairs, and $\circ$ is the warping operator. ${\mathcal L}_{S}$ represents the similarity (data) loss, while ${\mathcal L}_{R}$ denotes the regularization loss. The hyperparameter $\lambda$ balances the two terms. In experiments, we explored the use of MSE and NCC for ${\mathcal L}_{S}$, along with a first-order diffusion regularizer for ${\mathcal L}_{R}$.

On the other hand, LessNet can be directly applied to diffeomorphic image registration with only minor modifications to the network output and loss function. Computing a diffeomorphic deformation can be viewed as modeling a dynamical system \cite{beg2005computing} given by an ordinary differential equation (ODE): $\partial{\bf{\phi}}/\partial t  = {\boldsymbol{v}}_t({\bf{\phi}}_t)$, where ${\bf{\phi}}_0 = {\rm{Id}}$ represents the identity transformation, and ${\boldsymbol{v}}_t$ signifies the velocity field at time $t$ $(\in [0,1])$. Alternatively, a diffeomorphic deformation can be modeled with the stationary velocity field \cite{ashburner2007fast} through: $\partial{\bf{\phi}}/\partial t  = {\boldsymbol{v}}({\bf{\phi}}_t)$, where the velocity field ${\boldsymbol{v}}$ is assumed constant over time. In this paper, we utilized the stationary velocity field parameterized implementation. In such a case, the loss function of LessNet for diffeomorphic registration becomes:
\begin{equation}\label{eq:2}
   \argmin_{\boldsymbol{\Theta}}  \frac{1}{N} \sum_{i=1}^N  {\mathcal L}_{S}( I_{M_i} \circ  {\boldsymbol{Exp}}(\boldsymbol{v}_i),I_{F_i} )  + \frac{\lambda}{N} \sum_{i=1}^N   {\mathcal L}_{R}(\boldsymbol{v}_i ),
\end{equation}
with 
\begin{equation} \nonumber
    {\boldsymbol{v}_i} = {f}(\boldsymbol{\Theta}; I_{M_i}, I_{F_i}).
\end{equation}
The differences between the two losses \eqref{eq:1} and \eqref{eq:2} are twofold. First, \eqref{eq:2} requires the network $f$ to predict a stationary velocity field $\boldsymbol{v}$ rather than a displacement field $\boldsymbol{u}$. Second, in \eqref{eq:2}, we need to exponentiate the predicted stationary velocity field (i.e., ${\boldsymbol{Exp}}(\boldsymbol{v}))$ to attain a diffeomorphic deformation. To implement such an exponential function, we use seven scaling and squaring operations as in \cite{ashburner2007fast,dalca2018unsupervised}. All operations in LessNet are differentiable, thereby enabling optimization through standard backpropagation.

\section{Experiments}
\label{sec:exp}
\subsection{Datasets and Evaluation}

\textbf{OASIS-1 dataset \cite{marcus2007open}}: We used a processed version \cite{hoopes2021hypermorph} of this dataset, provided by the Learn2Reg challenge \footnote{\url{https://learn2reg.grand-challenge.org/Learn2Reg2021/}}, to perform subject-to-subject (inter-subject) brain registration. The dataset has 414 2D $160\times192$ slices and masks contain 24 anatomical structures extracted from their corresponding 3D $160\times192\times224$ volumes. We randomly split this 2D dataset into 201, 12, and 201 images for training, validation, and testing. After pairing, we ended up with 40200 ([201-1]$\times$201), 22 ([12-1]$\times$2), and 400 ([201-1]$\times$2) image pairs for training, validation, and testing, respectively.

\textbf{IXI dataset\footnote{\url{https://brain-development.org/ixi-dataset/}}}: We used a processed version of this dataset, provided by \cite{chen2022transmorph}, to perform atlas-to-subject registration. Following their exact evaluation protocol, we employed 403 images for training, 58 for validation, and 115 for testing. The atlas was generated by \cite{chen2022transmorph} with the method presented in \cite{kim2021cyclemorph}. All volumes and label maps (depicting 30 anatomical structures) were cropped to the size of 160$\times$192$\times$224.

{\textbf{Evaluation metrics}}: Dice score was employed to assess the overlapping ratio between anatomical structures. 
The better the registration performs, the higher the score. In addition, the percentage of negative Jacobian determinants in the deformation, denoted as $|J|_{< 0}\%$, was reported to assess whether the deformation is realistic and plausible. A higher percentage indicates more foldings in the deformation, while a lower percentage indicates fewer foldings. In the case of a perfect diffeomorphism, there will be no folding, and therefore the $|J|_{< 0}\%$ is expected to be zero.

\subsection{Implementation Details}

As shown in Fig. \ref{fig:LessNetArchitecture}, the overall model size of LessNet is controlled by the hyperparameter $C$, which we set to 8 by default. We implemented our proposed networks in PyTorch, where training was optimized using Adam with a learning rate of 0.0001 and a batch size of 1. To adapt LessNet to OASIS-1, we changed 3D kernels to 2D, each with a size of $3 \times 3$. For training in both 2D and 3D, we tuned built-in hyperparameters on respective held-out validation sets. In terms of loss functions, we employed MSE to train our networks on OASIS-1 for 20 epochs, achieving optimal performance with $\lambda=0.01$. On IXI, LessNet was trained with NCC for 500 epochs with $\lambda=5$, while the diffeomorphic LessNet (Diff-LessNet) was trained optimally with $\lambda=2$. We note that the optimal $\lambda$ was selected based on the highest Dice score obtained during tuning on the validation set. The training time depicted in Fig. \ref{fig:validation_accuracy_hours} was calculated using an A100 GPU. The CPU and GPU runtimes presented in Tables \ref{tab:compare_sota_oasis2d} and \ref{tab:ixi} were obtained from one local machine equipped with an RTX 2080Ti GPU, a 3.80GHz Intel(R) Core(TM) i7-9800X CPU with 128GB of RAM. The computational time, including the loading cost of models and images, was averaged over the entire test set with a batch size of 1.

\subsection{Impact of Pooling Features}

\begin{table}[t]
\caption{The impact of using multi-scale pooling features on the OASIS-1 dataset. For example, $\frac{1}{8}$ denotes the utilization of $6 \times \frac{H}{8} \times \frac{W}{8}$ pooling features, while $I_M$ and $I_F$ represent the use of the original images.}
\centering
% \footnotesize
\label{tab:ablation_level_pooling_features}
% \resizebox{0.84\linewidth}{!}{
% \setlength{\tabcolsep}{1.8mm}{
\begin{tabular}{lccccc}\hline
$\frac{1}{8} $     & $\frac{1}{4}$ & $\frac{1}{2}$ & $I_M , I_F$& Dice$\uparrow$        & $|J|_{ < 0}\%$  \\\hline
% Initial       & - & - & 0.544$\pm$0.089 \\
\cmark &  \xmark & \xmark & \xmark & 0.706$\pm$0.039 & 0.487$\pm$0.363 \\
\cmark &  \cmark & \xmark & \xmark & 0.748$\pm$0.038 & 0.644$\pm$0.362 \\
\cmark &  \cmark & \cmark & \xmark & 0.759$\pm$0.040 & 0.783$\pm$0.391 \\
\cmark &  \cmark & \cmark & \cmark & 0.761$\pm$0.039 & 0.742$\pm$0.353 \\\hline
\end{tabular}
% }
% }
\end{table}

First, we investigated the necessity of concatenating different resolutions of pooling feature maps as well as the original image pair to different blocks of the decoder of LessNet. To explore this, we conducted experiments with four different settings: 1) We disabled the concatenation of the original image pair ($I_M$ and $I_F$) and the pooling features to the last three conv blocks, but only concatenated the $6 \times \frac{H}{8} \times \frac{W}{8}$ pooling features to the first block; 2) We disabled the concatenation of the original image pair and $6 \times \frac{H}{2} \times \frac{W}{2}$ pooling feature, but concatenated the $6 \times \frac{H}{8} \times \frac{W}{8}$ and $6 \times \frac{H}{4} \times \frac{W}{4}$ pooling features; 3) We disabled the concatenation of only the original image pair, but concatenated the three different resolutions of pooling features; and 4) we enabled all the concatenations. 

As seen in Table \ref{tab:ablation_level_pooling_features}, the model achieved a Dice score of 0.706 under Setting 1. Registration performance got improved with the inclusion of more pooling features. In particular, there was a rapid increase in Dice to 0.748 under Setting 2, yielding a 4.2\% performance gain. A further increase of 1.1\% can be observed, achieving a Dice score of 0.759 under Setting 3. Finally, the highest performance was achieved under Setting 4, resulting in a Dice score of 0.761. These results underscore the necessity of utilizing multi-scale pooling features.

Next, we assessed the effectiveness of employing three different types of pooling operations: min pooling, max pooling, and average pooling. In Table \ref{tab:ablation_type_pooling_features}, the registration performances of min pooling and average pooling are comparable, achieving Dice scores of 0.757 and 0.758, respectively. Max pooling slightly lags behind at 0.753. Combining all three pooling operations resulted in the highest Dice score of 0.761.

\begin{table}[t]
\caption{Comparison of registration performance between different types of pooling operations on OASIS-1.}
\centering
% \footnotesize
\label{tab:ablation_type_pooling_features}
% \resizebox{0.84\linewidth}{!}{
% \setlength{\tabcolsep}{1.8mm}{
\begin{tabular}{ccccc}\hline
Min  & Average  & Max  & Dice$\uparrow$        & $|J|_{ < 0}\%$  \\\hline
% Initial       & - & - & 0.544$\pm$0.089 \\
\cmark & \xmark & \xmark &  0.757$\pm$0.040& 0.868$\pm$0.393\\
\xmark & \cmark & \xmark &  0.758$\pm$0.040& 0.802$\pm$0.409\\
\xmark & \xmark & \cmark &  0.753$\pm$0.041& 0.856$\pm$0.401\\
\cmark & \cmark & \cmark & 0.761$\pm$0.039 & 0.742$\pm$0.353 \\\hline
\end{tabular}
\end{table}

\subsection{Model Size}

As shown in Fig. \ref{fig:LessNetArchitecture}, the overall model size of LessNet is controlled by the hyperparameter $C$. Accordingly, we investigated the performance of LessNet with various values of $C$, namely, 4, 6, 8, 12, and 16. The results in Table \ref{tab:ablation_model_size} indicate that increasing the value of $C$ consistently improves registration performance, albeit at the cost of additional computational resources. For instance, when $C$ was set to 4, LessNet achieved a Dice score of 0.749, with 44,336 network parameters, 10.31MB of GPU memory with one forward and backward pass, and 152 million mult-adds. The Dice score increases by 0.8\%, 1.2\%, 1.7\%, and 1.9\%, respectively when $C$ was set to be 6, 8, 12, and 16. However, setting $C$ to 16 resulted in a 14.74-fold increase in parameters, a 3.93-fold increase in memory usage, and a 14.47-fold increase in mult-adds. 

\begin{table}[t]
\caption{Comparison between different LessNet architectures, with model sizes controlled by the hyperparameter $C$.}
\centering
% \footnotesize
\label{tab:ablation_model_size}
% \resizebox{0.84\linewidth}{!}{
% \setlength{\tabcolsep}{1.8mm}{
\begin{tabular}{cccccc}\hline
C  & Parameter  & Mult-Adds & Memory  & Dice$\uparrow$        & $|J|_{ < 0}\%$  \\\hline
% Initial       & - & - & 0.544$\pm$0.089 \\
4 & 44,336 & 152 & 10.31 & 0.749$\pm$0.040 & 0.808$\pm$0.389 \\
6 & 96,264 & 327 & 15.23 & 0.757$\pm$0.040 & 0.749$\pm$0.376 \\
8 & 168,032 & 570 & 20.16 & 0.761$\pm$0.039 & 0.742$\pm$0.353 \\
12 & 371,088  & 1250 & 30.00 & 0.766$\pm$0.039 &  0.852$\pm$0.392 \\
16 & 653,504  & 2200 &  39.84 & 0.768$\pm$0.039 & 0.830$\pm$0.397 \\\hline
\end{tabular}
% }
% }
\end{table}

\subsection{Diffeomorphism}

Finally, we investigated whether the proposed LessNet supports diffeomorphic registration. In this case, the output from the last convolutional layer was a stationary velocity field, and the diffeomorphic deformation field was integrated through 7 scaling and squaring operations \cite{dalca2018unsupervised, jia2022u, jia2023fourier}.

The results were presented in Table \ref{tab:ablation_diffeomorphism}, where we observed that LessNet can perform diffeomorphic registration well, as evidenced by nearly zero percentage of negative Jacobian determinants of the deformation. However, on this specific dataset, the registration performance of different diffeomorphic LessNets consistently lags behind their non-diffeomorphic counterparts in terms of Dice score. It is important to note that this phenomenon may not be universally valid for other datasets. For instance, diffeomorphic LessNet (Diff-LessNet) achieved higher accuracy than LessNet on IXI, as shown in Table \ref{tab:ixi}.

\begin{table}[t!]
\caption{Comparison of registration performance between diffeomorphic and non-diffeomorphic versions of LessNet.}
\centering
% \footnotesize
\label{tab:ablation_diffeomorphism}
% \resizebox{0.84\linewidth}{!}{
% \setlength{\tabcolsep}{1.8mm}{
\begin{tabular}{cccc}\hline
 Scaling \& Squaring  & $C$  & Dice$\uparrow$        & $|J|_{ < 0}\%$  \\\hline
% Initial       & - & - & 0.544$\pm$0.089 \\

\xmark &4& 0.749$\pm$0.040 & 0.808$\pm$0.389 \\
\cmark &4& 0.747$\pm$0.039 & $<$0.0001 \\
\xmark &6& 0.757$\pm$0.040 & 0.749$\pm$0.376 \\
\cmark &6& 0.754$\pm$0.039 & $<$0.0001 \\
\xmark &8& 0.761$\pm$0.039 & 0.742$\pm$0.353 \\
\cmark &8& 0.758$\pm$0.038 & $<$0.0001 \\
\xmark &12& 0.766$\pm$0.039 & 0.852$\pm$0.392 \\ 
\cmark &12& 0.761$\pm$0.038  &  $<$0.0001 \\
\xmark &16& 0.768$\pm$0.039 & 0.830$\pm$0.397 \\ 
\cmark &16& 0.762$\pm$0.039  &  $<$0.0001 \\\hline

\end{tabular}
\end{table}

\begin{table*}[ht]
\caption{Comparison of registration performance between various state-of-the-art methods on the OASIS-1 dataset.}
\centering
\footnotesize
\label{tab:compare_sota_oasis2d}
% \resizebox{0.84\linewidth}{!}{
\setlength{\tabcolsep}{1.8mm}{
\begin{tabular}{lccccccc}\hline
Methods & Dice$\uparrow$        & $|J|_{ < 0}\%$  & Parameters   & Mult-Adds (M) & Memory (MB)   & CPU (s) & GPU (s) \\\hline
Initial       & - & 0.544$\pm$0.089 &-           &-&-& - & -      \\
% Flash \cite{zhang2019fast}      & 16$\times$16      & 0.702$\pm$0.051 & 0.033$\pm$.126  & - & - & - & 13.699 & - \\
% Flash \cite{zhang2019fast}     & $20\times 24$      & 0.727$\pm$0.046 & 0.205$\pm$.279 & - & - & - &  22.575 & -    \\
Flash \cite{zhang2019fast}      & 0.734$\pm$0.045 & 0.049$\pm$0.080 & - & -& - & 85.773 & -    \\ \hline
VoxelMorph-1 \cite{balakrishnan2018unsupervised}  & 0.757$\pm$0.039 & 0.723$\pm$0.370 & 91,578 & 538.77& 15.29 & 0.007 & 0.014\\
VoxelMorph-2 \cite{balakrishnan2019voxelmorph}  & 0.757$\pm$0.040 & 0.793$\pm$0.405 & 100,530 & 690.29& 19.51 & 0.009 & 0.014 \\
Diff-VoxelMorph \cite{dalca2018unsupervised} & 0.740$\pm$0.044 & 0.019$\pm$0.082 & 102,532 & 294.28& 9.90 & 0.008 & 0.015  \\
TransMorph \cite{chen2022transmorph}   & 0.768$\pm$0.039 & 0.777$\pm$0.398 & 31,005,506
 & 3000& 30.53 & 0.036 & 0.021  \\
Diff-TransMorph \cite{chen2022transmorph}   & 0.748$\pm$0.043 & 0.022$\pm$0.076 & 30,934,084
 & 2170& 15.29 & 0.034 & 0.021  \\ \hline
% DeepFlash  \cite{wang2020deepflash} & 16$\times$16      & 0.615$\pm$0.055 & 0.0$\pm$0.0            & - & - & -& 0.487      &-   \\
% DeepFlash  \cite{wang2020deepflash} & $20\times 24$      & 0.597$\pm$0.066            & 0.0$\pm$0.0 & - & - & -&0.617    &-     \\
% DeepFlash   & 40x48      &             &             &     &     \\
% Diff-B-Spline \cite{qiu2021learning}  & $20\times 24$      & 0.710$\pm$0.041 &  0.014$\pm$0.072 & 70,226 & 86.16 & 6.58   &0.012     &0.015     \\
Diff-B-Spline \cite{qiu2021learning} & 0.737$\pm$0.038  & 0.015$\pm$0.069 & 88,690    & 139.40 & 7.49    &0.012 &0.015 \\
% \hline
% Fourier-Net$_{\text{Small}}$  & 0.748$\pm$0.039  & 0.672$\pm$0.391   &357,408          & 224.55 & 17.96    & {0.007}     & {0.014} \\
% Diff-Fourier-Net$_{\text{Small}}$  & 0.750$\pm$0.038  & $<$0.0001         &357,408    & 224.55 & 17.96    & 0.010     & 0.014 \\  
Fourier-Net \cite{jia2023fourier}  & 0.756$\pm$0.039  & 0.753$\pm$0.408       &1,427,376     & 888.25 & 35.89     & 0.011     & 0.015    \\
Diff-Fourier-Net \cite{jia2023fourier}  & 0.756$\pm$0.037  & $<$0.0001 &1,427,376 & 888.25 & 35.89  & 0.015     & 0.015   \\
% Fourier-Net${_{Large}}$  & 0.759$\pm$0.040  & 0.781$\pm$0.405             & 0.131     & 0.023    \\
% Diff-Fourier-Net${_{Large}}$ & \textbf{0.761$\pm$0.037}  & 0.0$\pm$0.0   & 0.040     & 0.024   \\\hline
% Fourier-Net+       & $40\times 48$   & 0.738$\pm$0.041  & 0.674$\pm$0.377 &300,477 & 142.66         & 9.52 & 0.006 & 0.014\\
% Diff-Fourier-Net+  & $40\times 48$   & 0.740$\pm$0.039  & 0.0$\pm$0.0    &300,477 & 142.66         & 9.52 & 0.009 & 0.014\\
% $4\times$Fourier-Net+$_{\text{Small}}$ & \textcolor{blue}{0.757$\pm$0.039}  & 0.252$\pm$0.197 &301,716 & 145.44         & 19.12 & 0.018 & 0.016\\
% Diff-$4\times$Fourier-Net+$_{\text{Small}}$ &0.746$\pm$0.040  & 0.0$\pm$0.0      &301,716 & 145.44         & 19.12 & 0.021 & 0.017\\
% $4\times$Fourier-Net+ & 0.761$\pm$0.039  & 0.278$\pm$0.232 &1,201,908 & 570.64        & 38.08 & 0.021 & 0.017\\
% Diff-$4\times$Fourier-Net+ &0.755$\pm$0.039  & $<$0.0001     &1,201,908  & 570.64         & 38.08 & 0.025 & 0.017\\
\hline
LessNet$_4$ & 0.749$\pm$0.040 & 0.808$\pm$0.389 & 44,336 & 151.80 & 10.31 & 0.006     & 0.014\\
Diff-LessNet$_4$ & 0.747$\pm$0.039 & $<$0.0001 & 44,336 & 151.80 & 10.31 & 0.009     & 0.014\\
LessNet$_6$ & 0.757$\pm$0.040 & 0.749$\pm$0.376 & 96,264 & 327.45 & 15.23 & 0.007     & 0.014\\
Diff-LessNet$_6$ & 0.754$\pm$0.039 & $<$0.0001 & 96,264 & 327.45 & 15.23 & 0.010     & 0.015\\
LessNet$_8$ & 0.761$\pm$0.039 & 0.742$\pm$0.353 & 168,032 & 569.59 & 20.16 & 0.008     & 0.014\\
Diff-LessNet$_8$ & 0.758$\pm$0.039 & $<$0.0001 & 168,032 & 569.59 & 20.16 & 0.012     & 0.015\\
LessNet$_{16}$ & 0.768$\pm$0.039 & 0.830$\pm$0.397 & 653,504 & 2200 & 39.84 & 0.015     & 0.015\\
Diff-LessNet$_{16}$& 0.762$\pm$0.038  &  $<$0.0001 & 653,504 & 2200 & 39.84 & 0.019     & 0.015
\\\hline
\end{tabular}
}
% }
\end{table*}

\begin{table*}[t!]
\centering
\footnotesize
\caption{Comparison of registration performance between different state-of-the-art methods on the 3D IXI dataset.}

% \resizebox{0.75\textwidth}{!}{
\begin{tabular}{lccccccc}
\hline
Methods            & Dice$\uparrow$ & $|J|_{ < 0}\%$     & Parameters & Mult-Adds (G) & Memory & CPU (s) & GPU (s)\\\hline
Affine    & 0.386$\pm$0.195 &- & -                      &-&-&-\\
SyN \cite{avants_ANTS}       & 0.645$\pm$0.152 & \textless{}0.0001      &-&-&-&-\\
NiftyReg \cite{modat2010fast} & 0.645$\pm$0.167 & \textless{}0.0001      &-&-&-&-\\
LDDMM \cite{beg2005computing}     & 0.680$\pm$0.135 & \textless{}0.0001      &-&-&-&-\\
Flash \cite{zhang2019fast}           & 0.692$\pm$0.140 & 0.0$\pm$0.0            &-&-&- & 1760& -\\
deedsBCV \cite{heinrich2015multi}  & 0.733$\pm$0.126 & 0.147$\pm$0.050        &-&-&-&-\\\hline
VoxelMorph-1 \cite{balakrishnan2019voxelmorph}  & 0.728$\pm$0.129 & 1.590$\pm$0.339    &274,387&304.05&2999.88&2.075&0.398\\
VoxelMorph-2 \cite{balakrishnan2019voxelmorph}  & 0.732$\pm$0.123 & 1.522$\pm$0.336    &301,411&398.81&3892.38&2.321&0.408\\
Diff-VoxelMorph \cite{dalca2018unsupervised}  & 0.580$\pm$0.165 & \textless{}0.0001      &307,878&89.67 &1464.26 &1.422 & 0.398\\
TransMorph \cite{chen2022transmorph}       & 0.754$\pm$0.124 & 1.579$\pm$0.328        &46,771,251&657.64 & 4090.31 & 4.094 & 0.516\\
% TransMorph-Bayes & 0.746$\pm$0.123 & & 1.560$\pm$0.333        &21,205,491&657.69\\
Diff-TransMorph \cite{chen2022transmorph}  & 0.594$\pm$0.163 & \textless{}0.0001      &46,557,414&252.61& 1033.18 &2.797  &0.419\\
Diff-B-Spline-TransMorph \cite{chen2022transmorph} & 0.761$\pm$0.122 & \textless{}0.0001      &46,806,307&425.95 & 1563.41& 7.582 & 0.417\\
LKU-Net \cite{jia2022u} & 0.765$\pm$0.129 & 0.109$\pm$0.054       &2,086,342&272.09 & 8713.36& 2.304 & 0.398\\
Diff-LKU-Net \cite{jia2022u} & 0.760$\pm$0.132 & 0.0$\pm$0.0      &2,086,342&272.09 & 8713.36& 5.914 & 0.390\\  \hline
Diff-B-Spline \cite{qiu2021learning}    & 0.742$\pm$0.128 & \textless{}0.0001      &266,387 & 47.05&1233.23 &5.649 & 0.378\\%\hline
% Fourier-Net$_{\text{Small}}$      & 0.759$\pm$0.132 & 0.009$\pm$0.008       &1,050,800&  43.82 & 2402.70 &{0.586} &\textbf{0.382}\\
% Diff-Fourier-Net$_{\text{Small}}$ & 0.756$\pm$0.130 & 0.0$\pm$0.0      &1,050,800&  43.82 & 2402.70 &4.164 &0.383\\
Fourier-Net \cite{jia2023fourier}      & {0.763$\pm$0.129} & 0.024$\pm$0.019       &4,198,352&  169.07 &4802.93& 1.029  &0.384\\
Diff-Fourier-Net \cite{jia2023fourier} & 0.761$\pm$0.131 & 0.0$\pm$0.0     &4,198,352&  169.07 &4802.93&4.668 &0.384\\
% Fourier-Net+       & 0.748$\pm$0.131  & 0.066$\pm$0.051 & 880,109& 19.30         & 670.20 & 0.455 & 0.381\\
% Diff-Fourier-Net+  & 0.750$\pm$0.130  & 0.0$\pm$0.0     & 880,109& 19.30         & 670.20& 4.035 & 0.378\\
% 3$\times$Fourier-Net+ & 0.766$\pm$0.131  & 0.015$\pm$0.011 & 2,640,327 & 57.90         & 2010.60 & 2.331 & 0.387 \\
% Diff-3$\times$Fourier-Net+ & 0.765$\pm$0.128  & 0.0$\pm$0.0     & 2,640,327 &57.90         & 2010.60 & 4.981 & 0.396 \\
\hline
LessNet$_4$ & 0.757$\pm$0.131  & 0.083$\pm$0.045     & 126,904 &60.26         & 1801.41 & 1.235 & 0.377 \\
Diff-LessNet$_4$ &  0.760$\pm$0.129 & 0.0$\pm$0.0     & 126,904 &60.26         & 1801.41 & 4.771 & 0.378 \\
LessNet$_6$ &   0.762$\pm$0.130  & 0.131$\pm$0.065     & 275,796 &127.97         & 2623.36 & 1.715 & 0.378 \\
Diff-LessNet$_6$ & 0.765$\pm$0.128  & 0.0$\pm$0.0     & 275,796 &127.97         & 2623.36 & 5.319 & 0.385 \\
LessNet$_8$ &   0.765$\pm$0.129  & 0.148$\pm$0.076     & 481,648 &220.74         & 3445.31 & 2.124 & 0.385 \\
Diff-LessNet$_8$ & 0.767$\pm$0.128  & 0.0$\pm$0.0     & 481,648 &220.74         & 3445.31 & 5.690 & 0.387 \\
LessNet$_{12}$ &  0.768$\pm$0.130 & 0.219$\pm$0.101    & 1,064,232 &481.43         & 5089.22 & 3.511 & 0.386 \\
Diff-LessNet$_{12}$ &  0.768$\pm$0.127 & 0.0$\pm$0.0     & 1,064,232 &481.43         & 5089.22 & 7.054 & 0.391 \\
\hline
\end{tabular}
% }
\label{tab:ixi}
\end{table*}

\subsection{Comparison with State-of-the-Art}

\subsubsection{Candidate Methods for Comparsion} 
In this section, we compared the effectiveness of our LessNet with 10 related methods detailed in the following. We note that for those methods requiring training, to ensure a fair comparison we used the Adam optimizer for all networks, with a learning rate of 0.0001 and a batch size of 1. Other hyperparameters were optimally tuned using the validation set for each network.

\begin{itemize}
    \item VoxelMorph. Two different non-diffeomorphic versions, namely VoxelMorph-1 \cite{balakrishnan2018unsupervised} and VoxelMorph-2 \cite{balakrishnan2019voxelmorph}, along with one diffeomorphic version \cite{dalca2018unsupervised}, were adopted using the official implementations available at \footnote{\url{https://github.com/voxelmorph/voxelmorph}}. Their official implementations were adapted to 2D registration by simply changing all 3D convolutional kernels into 2D. 
    \item TransMorph \cite{chen2022transmorph}. We adopted the official release of both 2D and 3D implementations, available at \footnote{\url{https://github.com/junyuchen245/TransMorph_Transformer_for_Medical_Image_Registration}}.
    \item Diff-B-Spline \cite{qiu2021learning}. The official implementation of this diffeomorphic B-Spline network was adopted from \footnote{\url{https://github.com/qiuhuaqi/midir}}. Specifically, for our 2D experiments, the optimal control spacing was set to 4, while for our 3D experiments, the control spacing was set to 3. The 3D results were directly copied from \cite{chen2022transmorph} and \cite{jia2023fourier+}.
    \item Flash \cite{zhang2019fast}, LKU-Net\cite{jia2022u}, and Fourier-Net \cite{jia2023fourier}. The results were directly copied from our previous publications \cite{jia2022u, jia2023fourier, jia2023fourier+}, as we used the exact same experimental settings on the two datasets.
    \item SyN \cite{avants_ANTS}, NiftyReg \cite{modat2010fast}, LDDMM \cite{beg2005computing}, deedsBCV \cite{heinrich2015multi}. We directly copied the relevant results on the 3D IXI dataset from \cite{chen2022transmorph} since we followed the same experimental settings.
\end{itemize}

\begin{figure*}
  \centering
  % include first image
  \includegraphics[width=.98\linewidth]{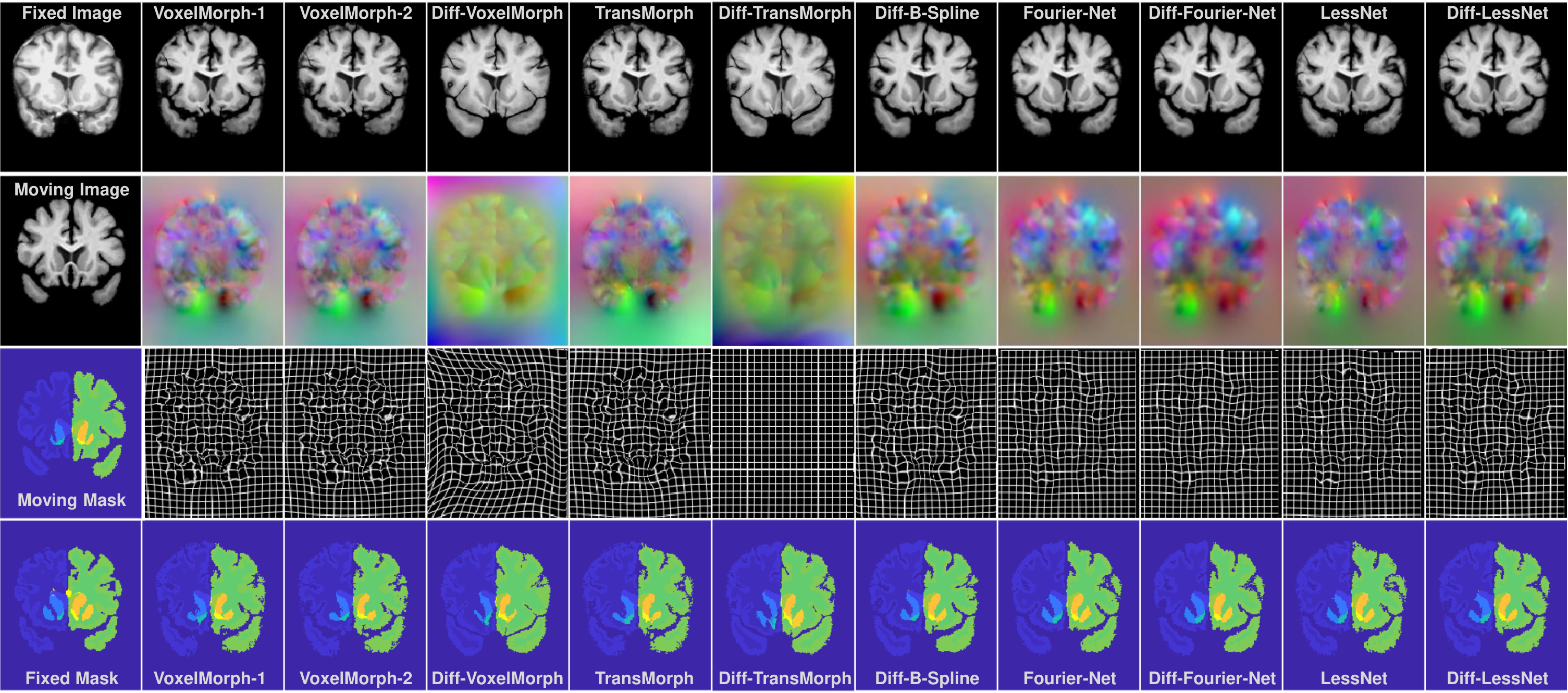}  
  \caption{Comparison of registration performance qualitatively. From top to bottom (apart from 1st column) are warped images, displacement fields, deformation fields, and warped moving masks.}
  
  \label{fig:visual_results_ixi}
\end{figure*}

\begin{figure*}[t]
\begin{subfigure}{.19\textwidth}
  \centering
  % include first image
  \includegraphics[width=.98\linewidth]{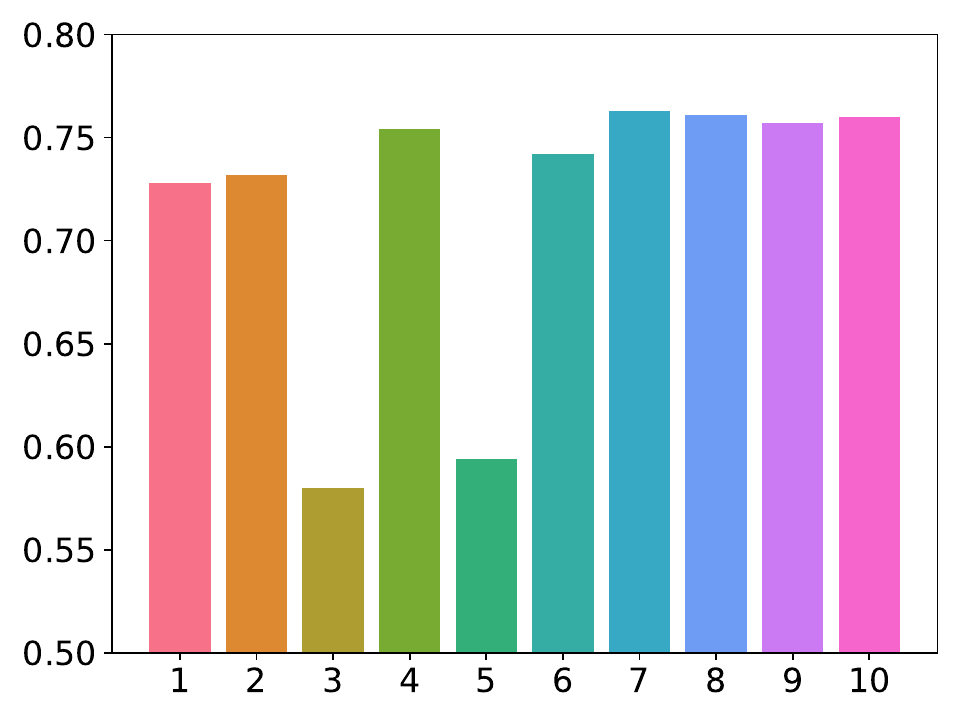} 
  \vspace{-18pt}
  \caption{Accuracy}
  \label{fig:sub-first}
\end{subfigure}
\begin{subfigure}{.19\textwidth}
  \centering
  % include second image
  \includegraphics[width=.98\linewidth]{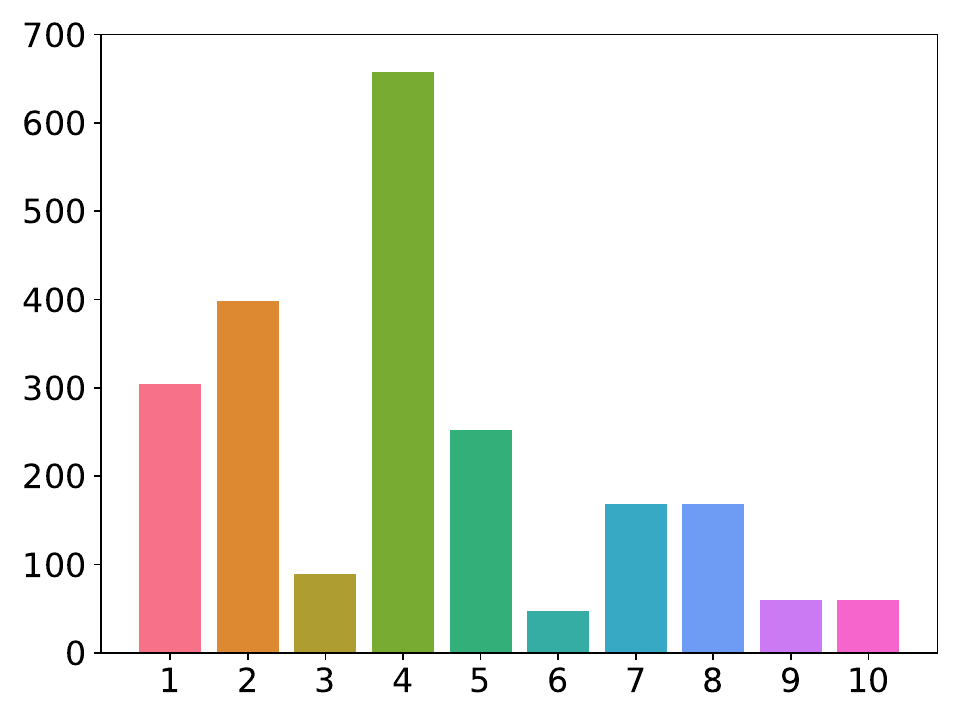}  
  \vspace{-18pt}
  \caption{Mult-Adds (G)}
  
  \label{fig:sub-second}
\end{subfigure}
\begin{subfigure}{.19\textwidth}
  \centering
  % include first image
  \includegraphics[width=.98\linewidth]{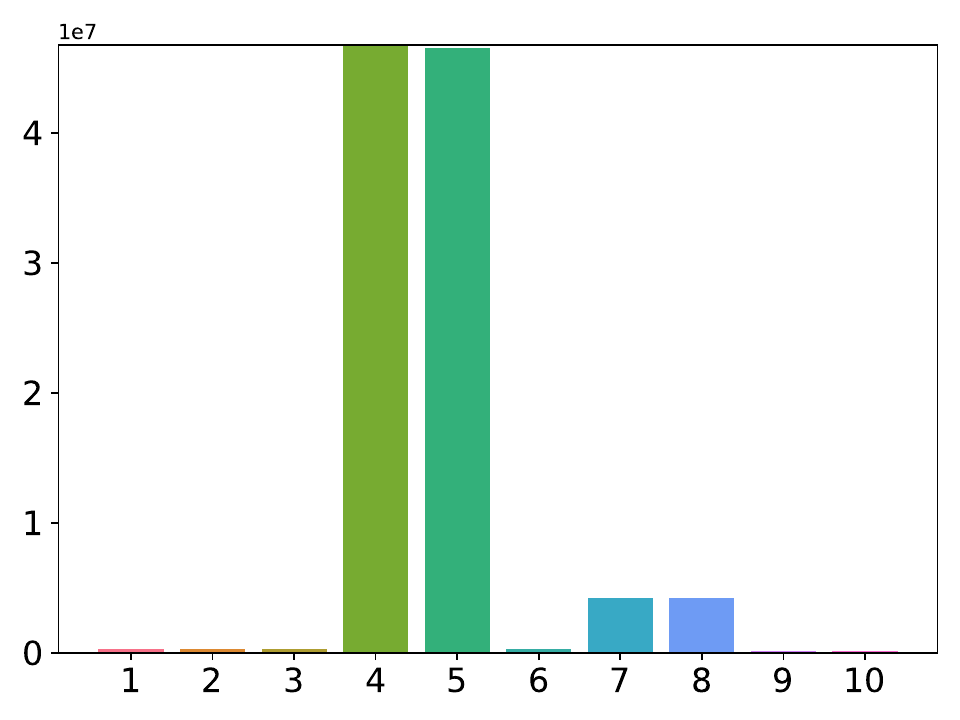}  
  \vspace{-18pt}
  \caption{Number of Parameters}
  \label{fig:sub-3rd}
\end{subfigure}
\begin{subfigure}{.19\textwidth}
  \centering
  % include second image
  \includegraphics[width=.98\linewidth]{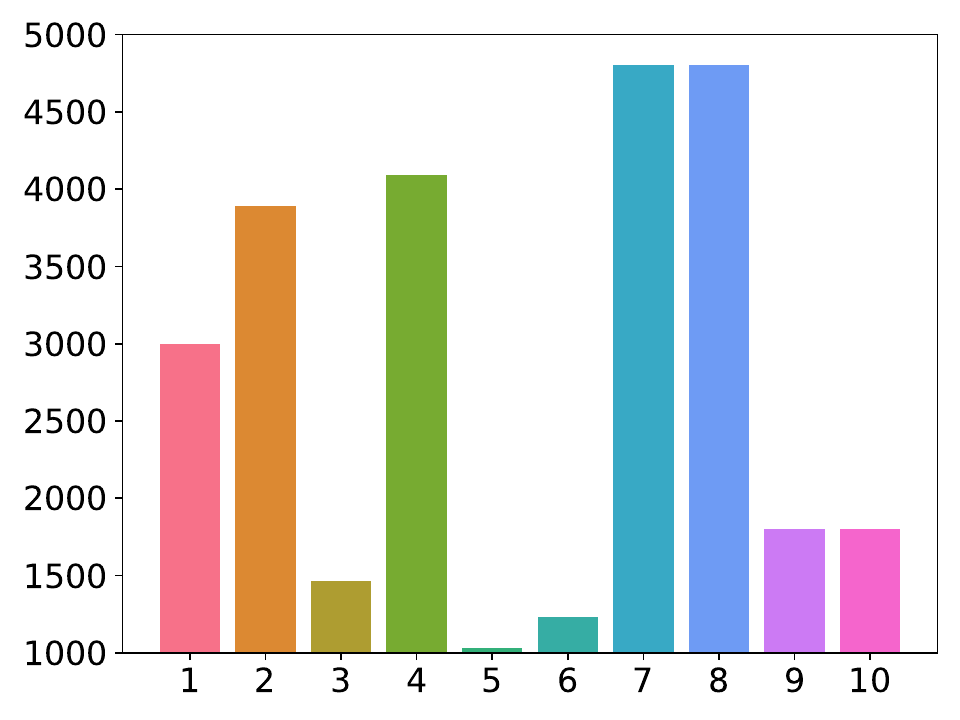}  
  \vspace{-18pt}
  \caption{GPU VRAM (GB)}
  \label{fig:sub-4th}
\end{subfigure}
\begin{subfigure}{.19\textwidth}
  \centering
  % include 6th image
  \includegraphics[width=.98\linewidth]{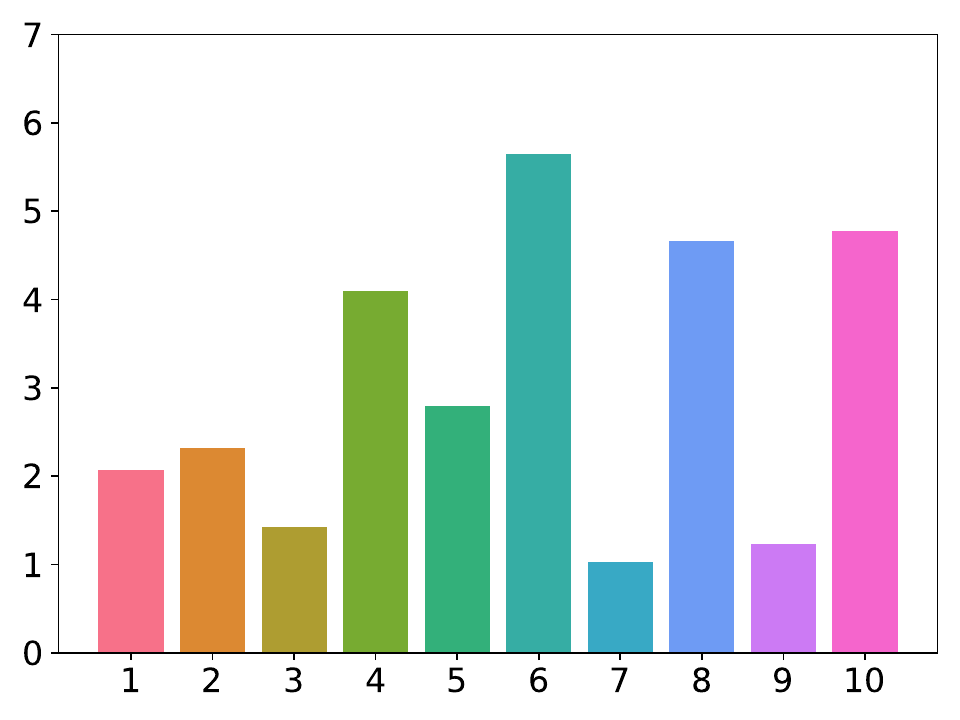}  
  \vspace{-18pt}
  \caption{CPU Inference Time (s)}
  \label{fig:sub-5th}
\end{subfigure}
\caption{Comparison between 10 different methods on various metrics such as registration accuracy, GPU memory usage, etc. From left to right in each plot: VoxelMorph-1, VoxelMorph-2, Diff-VoxelMorph, TransMorph, Diff-TransMorph, Diff-B-Spline, Fourier-Net, Diff-Fourier-Net, LessNet$_4$, Diff-LessNet$_4$.}
\label{fig:visual_barplot}
\vspace{-10pt}
\end{figure*}

\subsubsection{Subject-to-Subject Registration}
First, we note that we defined different versions of LessNet by appending a subscript denoting the parameter $C$. For example, LessNet$_4$ represents the architecture configured with $C$=4. As shown in Table \ref{tab:compare_sota_oasis2d}, with similar parameters and memory usage, our LessNet$_6$ has achieved comparable results to VoxelMorph-1 and VoxelMorph-2, using only 60.78\% and 47.44\% of their mult-adds. Furthermore, our LessNet$_{16}$ achieved comparable results with TransMorph while using only 2.11\% of its parameters and 73.33\% of its mult-adds. Moreover, our Diff-LessNet$_6$ outperformed both Diff-VoxelMorph and Diff-TransMorph by a large margin. These results show the superiority of LessNet over other encoder-decoder style networks.

On the other hand, our Diff-LessNet$_4$ is faster in CPU runtime and achieved a 1\% higher Dice score than Diff-B-Spline. Additionally, both our LessNet$_8$ and Diff-LessNet$_8$ outperformed Fourier-Net and Diff-Fourier-Net in terms of Dice and CPU runtime, further demonstrating its efficiency.

\subsubsection{Atlas-to-Subject  Registration}

Compared with encoder-decoder style networks that estimate dense full-resolution displacement fields, such as VoxelMorph-1, VoxelMorph-2, TransMorph, and LKU-Net, our LessNet achieved similar registration accuracy with lower computational cost and memory usage. Compared with model-driven networks that estimate low-dimensional representations of displacement fields, Fourier-Net+ and Diff-B-Spline demonstrated greater computational efficiency and lower memory usage than our LessNet. However, both our LessNet and Diff-LessNet exhibited comparable registration accuracy to these methods. To provide a comprehensive comparison, we have presented the qualitative registration results and quantitative plots of various methods in Figures \ref{fig:visual_results_ixi} and \ref{fig:visual_barplot}.

\section{Discussion and Conclusion}
In the context of unsupervised medical image registration using deep learning, there is a growing trend that larger encoder-decoder networks are proposed in pursuit of improved performance. Drawing inspiration from the classical VoxelMorph, we have identified redundancy within the encoder in this work. Building upon this insight, we introduced LessNet to prove that a learnable decoder alone can suffice for image registration. While we envision that LessNet can be extended to incorporate other handcrafted or pre-trained features, our current proof of concept focused solely on multi-scale pooling features. Our overall message from this paper is that for certain registration tasks, excessively large neural networks may not be imperative. Instead, the emphasis could be on designing more compact and efficient networks.

\bibliographystyle{IEEEtran}

\bibliography{IEEEabrv,ref}

\end{document}